\documentclass[letterpaper, 10 pt, conference]{ieeeconf}  

\IEEEoverridecommandlockouts                              

\usepackage{amsmath}
\usepackage{amssymb}  

\usepackage{todonotes}
\usepackage{cleveref}
\usepackage{pgfplots}
\usepackage{pgfplotstable}
\usepackage{filecontents}
\usepackage{tikz}
\usepackage{tikzscale}
\usepackage{tikz-3dplot}
\usepackage{multicol}

\usetikzlibrary{spy}
\usetikzlibrary{intersections}
\usetikzlibrary{arrows,shapes}
\usetikzlibrary{spy}
\usetikzlibrary{backgrounds}  
\usetikzlibrary{decorations}
\usetikzlibrary{decorations.markings}
\usetikzlibrary{positioning}
\usetikzlibrary{patterns}
\usetikzlibrary{calc} 
\usetikzlibrary{quotes} 
\usetikzlibrary{external} 
\usetikzlibrary{matrix}
\usepackage[inkscapearea=page,inkscapeversion=1]{svg}
\usepackage[normalem]{ulem}
\usepackage{subcaption}
\pgfplotsset{compat=1.17} 

\title{\LARGE \bf
Transfer Learning Enhanced Full Waveform Inversion*
}

\author{Stefan Kollmannsberger$^{1}$, Divya Singh$^{1}$ and Leon Herrmann$^{1}$
\thanks{The authors gratefully acknowledge the funding through the joint research project Geothermal-Alliance Bavaria (GAB) by the Bavarian State Ministry of Science and the Arts (StMWK) which finance Leon Herrmann as well as the Georg Nemetschek Institut (GNI) under the project Deep-Monitor which finances Divya Singh.}
\thanks{$^{1}$All authors are with the Technical University of Munich, Chair of Computational Modeling and Simulation, 80333 Munich, Bavaria, Germany (corresponding author: +49 89-289-25021; fax: +49 89-289-25051; e-mail: {\tt\small stefan.kollmannsberger@tum.de}}
}

\begin{document}

\maketitle
\thispagestyle{empty}
\pagestyle{empty}

\begin{abstract}

We propose a way to favorably employ neural networks in the field of non-destructive testing using Full Waveform Inversion (FWI). The presented methodology discretizes the unknown material distribution in the domain with a neural network within an adjoint optimization. To further increase efficiency of the FWI, pretrained neural networks are used to provide a good starting point for the inversion. This reduces the number of iterations in the Full Waveform Inversion for specific, yet generalizable settings.

\end{abstract}

\section{INTRODUCTION}
Full Waveform Inversion (FWI) is a well-known tool in non-destructive testing used to detect the exact location and shapes of flaws in a given artefact. To this end, FWI solves the following inverse-problem: Given a signal emitted at the boundary of a domain of interest as well as its measured image at selected locations, find the material distribution within the domain. The solution is found iteratively by computing the corresponding forward problem: given a domain with a material distribution and signal emitted at the boundary, compute the image of the emitted signal at selected locations. A review of FWI including an extension to the detection of voids can be found in~\cite{buerchner2022}.

Neural networks~\cite{Goodfellow-et-al-2016} are a promising tool for solving inverse problems \cite{kollmannsberger_deep_2021}. Two prominent methodologies exist. Firstly, the supervised learning method, where the mapping from the measured forward quantities to the inverse field is learned by interpolating labeled data. After training, cheap predictions can be made with unseen data. Applications of supervised learning in the context of ultrasonic non-destructive testing can e.g. be found in~\cite{alfarraj_petrophysical-property_2018, fabien-ouellet_seismic_2020, adler_deep_2019, yang_deep-learning_2019, wang_velocity_2020, wu_inversionet_2018, wang_velocity_2018, li_deep-learning_2020, zheng_applications_2019, araya-polo_deep_2019, mao_subsurface_2019, wu_seismic_2020, das_convolutional_2019, park_automatic_2020, araya-polo_deep-learning_2018, kim_geophysical_2018, Rao2021, Ye2022}. The second methodology are physics-informed neural networks (PINNs) \cite{psichogios_hybrid_1992, lagaris_artificial_1998, raissi_physics-informed_2019}, where one network learns the forward field (i.e. wave field traveling through the domain) while the second learns the inverse field (i.e. the field of the material coefficients). The networks are trained on the residual of the differential equation and a measurement residual. The state of the art concerning the application of classical PINNs and FWI is reported in~\cite{RashtBehesht2022}.

The supervised learning approach lacks both robustness and generalizability, while PINNs are not competitive compared to classical inversion frameworks, such as adjoint optimization~\cite{Fichtner2011}. However, reintroducing classical forward solvers while using a neural network as an Ansatz for the inverse field is beneficial since the neural network acts as a regularizer, as seen in~\cite{berg_neural_2021} for the Poisson equation, and in~\cite{xu_neural_2019} for the diffusion, wave and nonlinear Burger's equation. Compared to the classical FWI, this leads to smoother solutions without losing accuracy at material jumps i.e. flaws being located more accurately. Simultaneously, the convergence behavior and computational effort of the classical adjoint optimization are restored~\cite{herrmann2023}. 

In this contribution we will demonstrate that using the neural network as discretization of the inverse field, as presented in~\cite{herrmann2023}, offers the additional advantage of introducing transfer learning to speed up convergence. This combination can outperform classical inversion frameworks in terms of computational effort by reducing the number of iterations while maintaining the same quality of reconstruction. The basic idea of this combination is to pretrain the neural network on a labeled data set in an offline phase. It is then applied in an online phase to unseen inversion scenarios where it predicts a rough estimate of the spatial distribution of the field of the material coefficients. This provides a better starting point for the subsequent gradient-based inversion and thereby accelerates convergence. The pretraining is performed in a supervised manner. The influence of the size of the pretraining data set. Although investigations are conducted on the inverse problem of full waveform inversion, the framework is applicable to other types of inverse problems as well.

\section{Method}

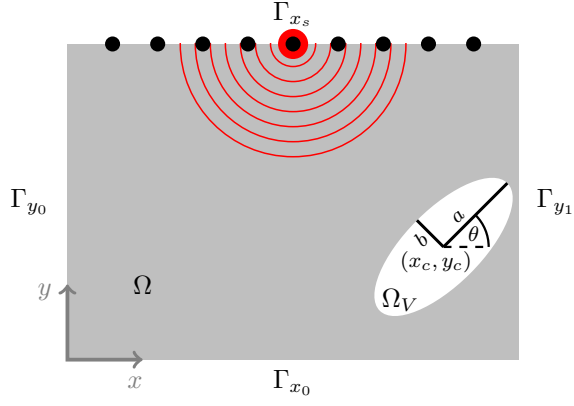
\begin{figure}[htb]
	\centering
	\begin{tikzpicture}
	\def\xc{5}
	\def\yc{1.5}
    \def\a{1.2} 
    \def\b{0.5} 
    \def\angle{45} 
  
	\fill [lightgray, line width=0.3mm] (0,0) rectangle (6,4.2);
    \fill [line width=0.3mm, white, rotate around={\angle:({\xc},{\yc})}] ({\xc},{\yc}) ellipse ({\a} and {\b});
    
    \draw [line width=0.4mm, rotate around={\angle:(\xc, \yc)}] (\xc, \yc) -- (\xc+\a, \yc);
    \draw [line width=0.4mm, rotate around={\angle:(\xc, \yc)}] (\xc, \yc) -- (\xc, \yc+\b);
    \draw [line width=0.3mm, dashed] (\xc, \yc) -- (\xc+0.6, \yc);
    \draw [line width=0.3mm] (\xc+0.6, \yc) arc [radius=0.6, start angle=0, end angle=\angle];
    
    \node [rotate=\angle] at (\xc+0.2, \yc+0.4) {\footnotesize $a$};
    \node [rotate=\angle] at (\xc-0.3, \yc+0.1) {\footnotesize $b$};    
    \node at (\xc+0.4, \yc+0.17) {\footnotesize $\theta$};
    \node at (\xc-0.1, \yc-0.2) {\footnotesize $(x_c, y_c)$};
    

	\node [left] at (\xc-0.2,\yc-0.7) {$\Omega_V$};
	
	\begin{scope}
	\clip (0,0) rectangle (6,4.2);
	\foreach {\r} in {0.3,0.5,...,1.5} {
		\draw [line width=0.2mm, red] (3,4.2) circle (\r cm);
	}
	\end{scope}

	\draw [gray, line width=0.5mm,<->] (0,1) -- (0,0) -- (1,0);
	\node [gray] at (-0.3,0.9) {$y$};
	\node [gray] at (0.9, -0.3) {$x$};

	\fill [red] (3,4.2) circle (0.2cm);

	\foreach {\x} in {0.6,1.2,...,5.4} {
		\foreach {\y} in {4.2} {
			\fill [black] (\x,\y) circle (0.1cm);
		}
	}

	\node at (-0.5,2.1) {$\Gamma_{y_0}$};
	\node at (6.5,2.1) {$\Gamma_{y_1}$};
	\node at (3,-0.3) {$\Gamma_{x_0}$};
	\node at (3,4.6) {$\Gamma_{x_s}$};
	\node at (1,1) {$\Omega$};
	\end{tikzpicture}

	\caption{Ultrasonic Nondestructive Testing on a two-dimensional rectangular domain $\Omega$ with a void $\Omega_V$. Horizontal boundaries are defined as $\Gamma_{x}=\Gamma_{x_0}\cup\Gamma_{x_1}$, while vertical boundaries are defined as $\Gamma_{y}=\Gamma_{y_0}\cup\Gamma_{y_1}$. The ellipsoidal defect is parametrized by the two half-axes $a, b$, the center $x_c, y_c$ and the rotation $\theta$.}
	\label{fig:ndt}
\end{figure}

The general setting of the FWI problem is given in~\cref{fig:ndt} which depicts a domain $\Omega$ with a homogeneous material distribution and a flaw in form of a void represented by an ellipse. The task is to find the location and geometry of that void by FWI. To this end, an emitter in red is inducing a signal $u(x_s,t)$ at a specific location $x_s$ at the boundary $\Gamma$ of a computational domain.
The signal is generated by a volume force $f(\boldsymbol{x},t)$ modeled by a spatial Dirac delta using a source term $\psi(t)$.
\begin{equation}
f(\boldsymbol{x},t)=\psi(t)\delta(\boldsymbol{x}-\boldsymbol{x}_s) \label{eq:source}
\end{equation}
A common choice for the source term is a sine burst with $n_c$ cycles, a frequency $\omega = 2 \pi f_{\psi}$, and an amplitude $\psi_0$, so that
\begin{equation}
\psi(t)=\begin{cases}
\psi_0 \sin(\omega t)\sin(\frac{\omega t}{2 n_c}), & \text{for } 0\leq t\leq \frac{2\pi n_c}{\omega} \\
0, & \text{for } \frac{2\pi n_c}{\omega}<t. \label{eq:sineburst}
\end{cases}
\end{equation}

It is now assumed that the signal $u(\boldsymbol{x},t)$ spreads throughout the medium characterized by the wave equation:
\begin{align}
\gamma(\boldsymbol{x}) \rho_0 u_{tt}(\boldsymbol{x}, t)-\nabla\cdot(\gamma(\boldsymbol{x}) \rho_0 c_0^2 \nabla u(\boldsymbol{x}, t)) = f(\boldsymbol{x},t)  \label{eq:waveequation}
\end{align}
defined in space and time $\Omega \times\mathcal{T}$. The medium is assumed to have a constant wave speed $c_0$ as well as a constant density $\rho_0$. \Cref{eq:waveequation} is further suppield with the typical boundary and initial conditions for the wave field:
\begin{alignat}{2}
u_x(\boldsymbol{x}, t)&=0  && \qquad \text{on }  \Gamma_{y}\times\mathcal{T} \label{eq:neumanninx} \\ 
u_y(\boldsymbol{x}, t)&=0  && \qquad \text{on } \Gamma_{x}\times\mathcal{T} \label{eq:neumanniny} \\
u(\boldsymbol{x}, 0)&=u_t(\boldsymbol{x}, 0)=0 && \qquad \text{on } \Omega. \label{eq:initialconditions} 
\end{alignat}
The symbols $u_{t}$ and $u_{tt}$ mark the first and the second time derivative of the signal. 

Most importantly, in~\cref{eq:waveequation} $\gamma(\boldsymbol{x})$ denotes the unknown which is to be determined in the process of FWI. The function $\gamma(\boldsymbol{x})$ is a scaling function which is constrained between $[\epsilon, 1]$, where $\epsilon \ll 1$ is a non-zero lower bound which indicates a local defect of the material in form of a void. 

In the sequel we assume that the unknown scaling function $\gamma(\boldsymbol{x})$ is predicted by a neural network $A_{\gamma}$ with trainable parameters $\boldsymbol{\theta}_{\gamma}$:
\begin{align}
\hat{\gamma}(\boldsymbol{\theta}; \boldsymbol{x})&=A_{\gamma}(\boldsymbol{\theta}; \boldsymbol{x}) \label{eq:coefficient}
\end{align}

The mathematical model of the wave equation given by~\cref{eq:waveequation} is now used together with~\cref{eq:coefficient} to predict a signal $\boldsymbol{\hat{u}}$ with the help of a deterministic operator $F$:
\begin{align}
\hat{u}(\hat{\gamma}(\boldsymbol{\theta}; \boldsymbol{x}), t)&=F(\hat{\gamma}(\boldsymbol{\theta}; \boldsymbol{x}), t)\label{eq:predu}
\end{align}
In this contribution, the deterministic operator $F$ is a finite difference disrectization in space and time. The prediction of the fields is denoted using the hat symbol over a variable. The indicator field over the entire domain $\Omega$ is denoted in bold as $\boldsymbol{\gamma}$. Accordingly, the wavefield over the domain spatio-temporal domain $\Omega\times\mathcal{T}$ is denoted as $\boldsymbol{u}$.

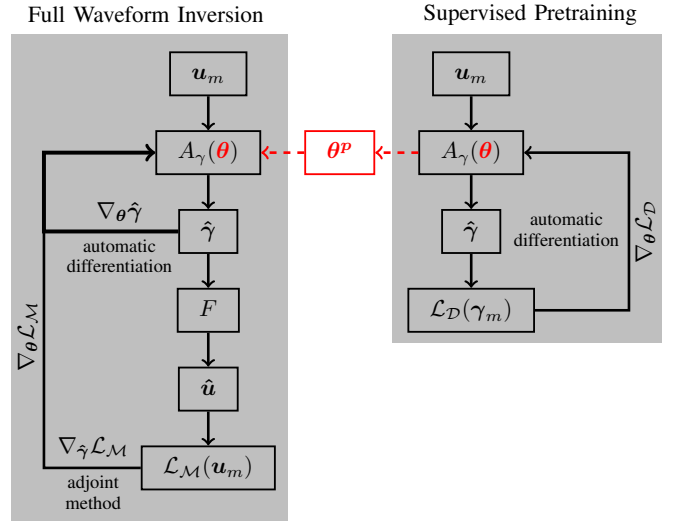
\begin{figure}[htb]
	\centering
	\resizebox{0.5\textwidth}{!}{
	\begin{tikzpicture}
	
	\fill [lightgray] (-3, -6.8) rectangle (1.2,0.6);
	\node at (-0.95,0.9) {Full Waveform Inversion};
	
	\node (I) [draw, thick] at (0,0) {\begin{tabular}{c} $\boldsymbol{u}_m$ \end{tabular}};
	\node (A) [draw, thick] at (0,-1.2) {\begin{tabular}{c} $A_{\gamma}({\color{red}\boldsymbol{\theta}})$ \end{tabular}};
	\node (G) [draw, thick] at (0,-2.4) {\begin{tabular}{c} $\boldsymbol{\hat{\gamma}}$ \end{tabular}};
	\node (F) [draw, thick] at (0,-3.6) {\begin{tabular}{c} $F$ \end{tabular}};
	\node (U) [draw, thick] at (0,-4.8) {\begin{tabular}{c} $\boldsymbol{\hat{u}}$ \end{tabular}};
	\node (L) [draw, thick] at (0,-6) {\begin{tabular}{c} $\mathcal{L}_{\mathcal{M}}(\boldsymbol{u}_m)$ \end{tabular}};	
	
	\draw [line width=0.4mm,,->] (I.south) -- (A.north);
	\draw [line width=0.4mm,,->] (A.south) -- (G.north);
	\draw [line width=0.4mm,,->] (G.south) -- (F.north);
	\draw [line width=0.4mm,,->] (F.south) -- (U.north);
	\draw [line width=0.4mm,,->] (U.south) -- (L.north);
	
	\draw [line width=0.4mm,,-] (L.west) -- (-2.5,-6) -- (-2.5,-2.4) -- (G.west);
	\draw [line width=0.6mm,,->] (G.west) -- (-2.5,-2.4) -- (-2.5,-1.2) -- (A.west);
	\node [black] at (-1.75,-5.7) {$\nabla_{\boldsymbol{\hat{\gamma}}}\mathcal{L}_\mathcal{M}$};
	\node [black] at (-1.75,-6.4) {\footnotesize \begin{tabular}{c}adjoint\\ method \end{tabular}};
	\node [black] at (-1.35,-2.1) {$\nabla_{\boldsymbol{\theta}}\boldsymbol{\hat{\gamma}}$};
	\node [black] at (-1.35,-2.8) {\footnotesize \begin{tabular}{c}automatic\\ differentiation \end{tabular}};
	\node [black, rotate=90] at (-2.75, -4) {$\nabla_{\boldsymbol{\theta}}\mathcal{L}_\mathcal{M}$};
	
	
	\fill [lightgray] (2.8, -4.1) rectangle (6.9,0.6);
	\node at (4.9,0.9) {Supervised Pretraining};
	
	\node (IS) [draw, thick] at (4,0) {\begin{tabular}{c} $\boldsymbol{u}_m$ \end{tabular}};
	\node (AS) [draw, thick] at (4,-1.2) {\begin{tabular}{c} $A_{\gamma}({\color{red}\boldsymbol{\theta}})$ \end{tabular}};
	\node (GS) [draw, thick] at (4,-2.4) {\begin{tabular}{c} $\boldsymbol{\hat{\gamma}}$ \end{tabular}};
	\node (LS) [draw, thick] at (4,-3.6) {\begin{tabular}{c} $\mathcal{L}_{\mathcal{D}}(\boldsymbol{\gamma}_m)$ \end{tabular}};	
	
	\node (TP) [draw, thick, red] at (2,-1.2) {\begin{tabular}{c} ${\color{red}\boldsymbol{\theta^p}}$ \end{tabular}};
	
	\draw [red, dashed, line width=0.4mm,->] (TP.west) -- (A.east);
	\draw [red, dashed, line width=0.4mm,->] (AS.west) -- (TP.east);
	\draw [line width=0.4mm,,->] (IS.south) -- (AS.north);
	\draw [line width=0.4mm,,->] (AS.south) -- (GS.north);
	\draw [line width=0.4mm,,->] (GS.south) -- (LS.north);
	\draw [line width=0.4mm,,->] (LS.east) -- (6.4,-3.6) -- (6.4,-1.2) -- (AS.east);
	
	\node [black] at (5.45,-2.4) {\footnotesize \begin{tabular}{c}automatic\\ differentiation \end{tabular}};
	\node [black, rotate=90] at (6.65, -2.4) {$\nabla_{\boldsymbol{\theta}}\mathcal{L}_\mathcal{D}$};
	
	\end{tikzpicture}
	}
	\caption{Schematic of the FWI training procedure on the left and the possible supervised pretraining on the right to accelerate the convergence with good starting weights $\boldsymbol{\theta}^p$. In the FWI, first the sensitivity of the cost function $\mathcal{L}$ w.r.t. the predicted indicator field $\boldsymbol{\hat{\gamma}}$ is computed with the adjoint method using~\cref{eq:frechet,eq:adjointsensitivity}. The sensitivity of the indicator field $\boldsymbol{\hat{\gamma}}$ w.r.t. the neural network parameters $\boldsymbol{\theta}$ is then found with automatic differentiation in the form of backpropagation frequently used  to train neural networks. This yields the desired gradient $\nabla_{\boldsymbol{\theta}}\mathcal{L}$ by using the chain rule~\cref{eq:chainrule}. The supervised pretraining is conducted in a conventional manner with labeled data as e.g. elaborated in~\cite{Goodfellow-et-al-2016}.}
	\label{fig:NewApproach}
\end{figure}

We now consider a virtual experiment with the same setting, where a known signal $u$ is emitted at the same locations $x_s$ as in the computational model used for the prediction of $\hat{\boldsymbol{\gamma}}$. The signal is recorded by the sensors illustrated by black circles in~\cref{fig:ndt} and called measurement $u_\mathcal{M}(x_i,t)$. Its deviation from the signal predicted by the mathematical model $\hat{u}$ is measured by the following loss functional, also referred to as the cost function:

\begin{align}
\begin{split}
\mathcal{L}_{\mathcal{M}}(\hat{u}(\boldsymbol{\hat{\gamma}}(\boldsymbol{\theta}))) 
 &= \\
 &\frac{1}{2}\int_\mathcal{T}\int_\Omega \sum_{i=1}^{N_{\mathcal{M}}} \biggl(\hat{u}\left(\hat{\gamma}(\boldsymbol{\theta}; \boldsymbol{x}), t\right) - u_{\mathcal{M}}(\boldsymbol{x}_i, t)\biggr)^2 \\
 &\delta(\boldsymbol{x} - \boldsymbol{x}_i) d\Omega d\mathcal{T}. \label{eq:measurementloss}
 \end{split}
\end{align}
The minimization of this cost function leads to the discovery of the scaling function $\boldsymbol{\hat{\gamma}}$ and thereby the spatial location of voids in the material\footnote{Note that the dependency of $\boldsymbol{\hat{\gamma}}$ on $\boldsymbol{x}$ and $\boldsymbol{\hat{u}}$ on $\boldsymbol{x}, t$ is not always explicitly given to declutter the notation.}. The minimization procedure is depicted on the left in~\cref{fig:NewApproach} and briefly explained in the sequel. 

The functional given in~\cref{eq:measurementloss} is minimized iteratively using a gradient based approach yielding an updated indicator field $\boldsymbol{\gamma}^{(j+1)}$ parametrized by the new weights $\boldsymbol{\theta}^{(j+1)}$.
\begin{equation}
\boldsymbol{\theta}^{(j+1)}=\boldsymbol{\theta}^{(j)}-\alpha \nabla_{\theta^{(j)}} \mathcal{L}_\mathcal{M}(\hat{u}(\boldsymbol{\hat{\gamma}}(\boldsymbol{\theta}^{(j)})))\label{eq:gradientDecent}
\end{equation}
Here, $\alpha$ is the step size, also referred to as learning rate, the superindex $(j)=1,2,\dots$ is the iteration counter. This iterative process then yields an optimized estimate of the scaling field $\boldsymbol{\hat{\gamma}}(\boldsymbol{\theta})$ which gives a direct indication of the flaw. 

\Cref{eq:gradientDecent} requires the computation of the gradient of the cost function with respect to the network parameters $\boldsymbol{\theta}$ and can efficiently be computed by using the chain rule:
\begin{equation}
\nabla_{\boldsymbol{\theta}}\mathcal{L}(\hat{u}(\boldsymbol{\hat{\gamma}}(\boldsymbol{\theta})))= \underbrace{\nabla_{\boldsymbol{\hat{\gamma}}}\mathcal{L}(\hat{u}(\boldsymbol{\hat{\gamma}}))}_{\text{\normalfont adjoint method}} \quad \cdot \underbrace{\nabla_{\boldsymbol{\theta}}\boldsymbol{\hat{\gamma}}(\boldsymbol{\theta})}_{\text{\normalfont automatic differentiation}} \label{eq:chainrule}
\end{equation}
The first term on the right hand side is the outer derivative i.e. the gradient of the cost function w.r.t. the indicator field $\nabla_{\boldsymbol{\hat{\gamma}}}\mathcal{L}$. It is efficiently computed with the continuous adjoint method~\cite{buerchner2022} using the Fr\'echet kernel $K_{\gamma}$:
\begin{equation}
K_{\gamma}=\int_{\mathcal{T}} \left[-\rho_0 u_t^{\dagger} u_t+\rho_0 c_0^2 \nabla u^{\dagger} \cdot \nabla u\right] dt \label{eq:frechet}
\end{equation}
which is integrated in space resulting in the gradient:
\begin{equation}
\nabla_{\boldsymbol{\hat{\gamma}}}\mathcal{L}(\hat{u}(\boldsymbol{\hat{\gamma}}))=\int_{\Omega} K_{\gamma} d\Omega \label{eq:adjointsensitivity}
\end{equation}
See~\cite{buerchner2022} for more details. To compute~\cref{eq:frechet}, the temporal and spatial gradients of the solution field $u$ and the adjoint field $u^{\dagger}$ are computed with finite differences. The second term in~\cref{eq:chainrule} is the inner derivative, i.e. the sensitivity of the scaling function w.r.t. the neural network parameters $\nabla_{\boldsymbol{\theta}}\boldsymbol{\hat{\gamma}}$. This derivative is obtained via automatic differentiation in the form of backpropagation that is frequently used  to train neural networks \cite{Goodfellow-et-al-2016}, in this case  $A_{\boldsymbol{{\gamma}}}(\boldsymbol{\theta})$.

The presented framework for the detection of flaws by using a scaling function discretized by neural networks as briefly recalled above was already explained in a more elaborate form in~\cite{herrmann2023}. Therein, it was shown to deliver more accurate results than classical methods. This can be attributed to the favourable properties of $A_\gamma$. Although the neural network as discretization adds complexity to the FWI optimization in terms of the required automatic differentiation, the additional cost is negligible next to the evaluation of the forward operator $F$ for a large number of timesteps. Thus similar computational times and number of iterations are observed as in classical FWI. Yet, the presented framework has many advantages, one of which is the possibility of reducing the number of required iterations and is elaborated next.

As in any gradient based approach, the choice of the starting point $\boldsymbol{\hat{\gamma}}^{(0)}$ decisively determines the number of necessary iterations. The closer this starting point is located to the optimal $\hat{\boldsymbol{\gamma}}$, the less iterations are needed. This observation offers the chance to exploit the concept of transfer learning \cite{pan_survey_2010} for a better starting point $\boldsymbol{\hat{\gamma}}^{(0)}$ obtained by better starting weights $\boldsymbol{\theta}^{(0)}$. The neural network is pretrained in a supervised manner \cite{Ye2022,Rao2021} with data pairs $(\boldsymbol{u}_\mathcal{M}, \boldsymbol{\gamma}_\mathcal{M})$ using a data-driven cost function defined in terms of the mean squared error
\begin{equation}
    \mathcal{L}_\mathcal{D}(\boldsymbol{\hat{\gamma}}(\boldsymbol{\theta}))=\frac{1}{2}\sum_{i=1}^{N_{\mathcal{D}}} \int_\Omega \biggl( \hat{\gamma}_i(\boldsymbol{\theta}; \boldsymbol{x}) - \gamma_{\mathcal{M}_i}(\boldsymbol{x}) \biggr)^2 d\Omega
    \label{eq:cost_func}
\end{equation}
The supervised pretraining depicted on the right in~\cref{fig:NewApproach} yields a good starting weight $\boldsymbol{\theta}$ to be saved as $\boldsymbol{\theta}^p$ and transferred via $A_\gamma$ to the FWI framework, where it provides a good approximation to $\hat{\boldsymbol{\gamma}}^{(0)}$ simply by evaluating the pretrained network $A_{\gamma}(\boldsymbol{\theta}^p)$: 
\begin{align}
\hat{\boldsymbol{\gamma}}^{(0)} &=A_{\gamma}(\boldsymbol{\theta}^p) \label{eq:pretrainedcoefficient}
\end{align}

The starting point $\boldsymbol{\hat{\gamma}}^{(0)}$ can then directly be used as the starting point for the iteration as depicted on the right in~\cref{fig:NewApproach}. This is possible due to the identical network architecture used in the supervised training and FWI in the approximation of the field  $\hat{\gamma}(\boldsymbol{\theta}; \boldsymbol{x})$ as given in~\cref{eq:coefficient}. This enables the direct use of the network parameters $\boldsymbol{\theta}^p$ as a starting point for the inversion, see~\cref{eq:gradientDecent} and~\cref{fig:NewApproach}. 

\begin{figure*}
  \includegraphics[width=\textwidth]{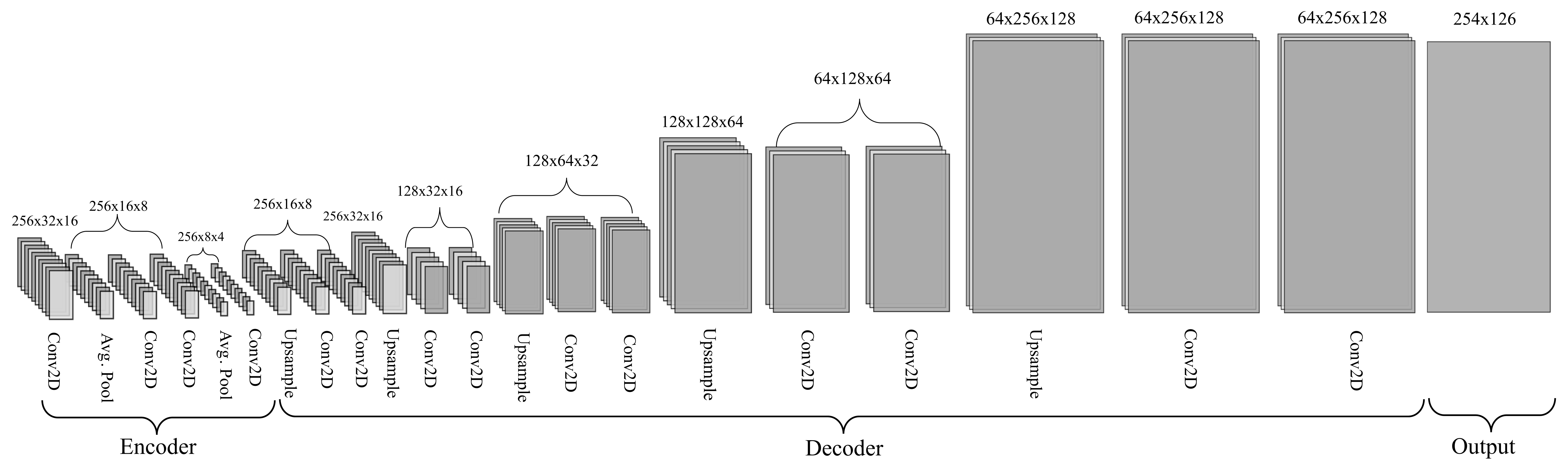}
  \caption{The encoder-decoder neural network architecture $A_\gamma$ discretizing the material $\boldsymbol{\hat{\gamma}}$ given a signal $\boldsymbol{u}_{\mathcal{M}}$.}
  \label{fig:CNN_arc}
\end{figure*}

\Cref{sec:results} investigates the improvement shown by a pretrained Neural network $A_{\gamma}$, for carrying out the FWI. For this purpose we use a network comprised of an encoder and a decoder which was first proposed in~\cite{kalchbrennerrecurrent, NIPS2014_a14ac55a} for language translation. The architecture of the neural network is shown in~\cref{fig:CNN_arc}. 
The input to this encoder-decoder network is chosen as the last quarter of the signal since it contains the most information about the damage present in the domain due to multiple reflections. This input data is first reshaped from a 1D vector of 512 data points to a 2D matrix of $32\times16$ before being fed to the neural network. 

The first half of the neural network consists of the encoder layers that are made up of a combination of convolution and average pooling. The activation function used for these layers is the PReLU function~\cite{he_delving_2015}. Followed by this, the decoder network scales up the compressed input received from the encoder with the help of a combination of convolutional and upsampling' layers (with nearest-neighbour interpolation to retain the discontinuities). The same activation function was used as before except for the last layer where an adaptive sigmoid function~\cite{jagtap_adaptive_2020} is used to ensure that the output is between 0 and 1. This was done in order to match the values of the field $\hat{\gamma}(\boldsymbol{\theta}; \boldsymbol{x})$ which are 1 at undamaged locations and 0 at the damaged location. The optimization is performed with the Adam~\cite{kingma2014adam} optimizer. The initial weights are set with Glorot initialization~\cite{kumar2017weight}. 

The dataset $(\boldsymbol{u}_\mathcal{M}, \boldsymbol{\gamma}_\mathcal{M})$ is generated with elliptical defects parametrized by the center $x_c, y_c$, the semi-major and semi-minor axes $a, b$, and the rotation $\theta$, as illustrated in~\cref{fig:ndt}. The wave measurements $\boldsymbol{u}_\mathcal{M}$ are then obtained by a forward solution with finite differences\footnote{Note, that this is an inverse crime~\cite{wirgin_inverse_2004}, as the reference data is generated with the same numerical method used in the FWI. For the sake of simplicity and to illustrate the approach, this is however tolerated and not deemed to be an obstacle to the proposed method.}. In total $2100$ samples are obtained and used throughout the upcoming studies. Up to $2048$ samples are used for pretraining, while the last $62$ are always unseen before application of the FWI. These are thus reserved as validation.

\section{Results} \label{sec:results}

\begin{figure}[htb]
\centering
\begin{subfigure}[b]{0.41\textwidth}
\centering
\includegraphics[width=\textwidth]{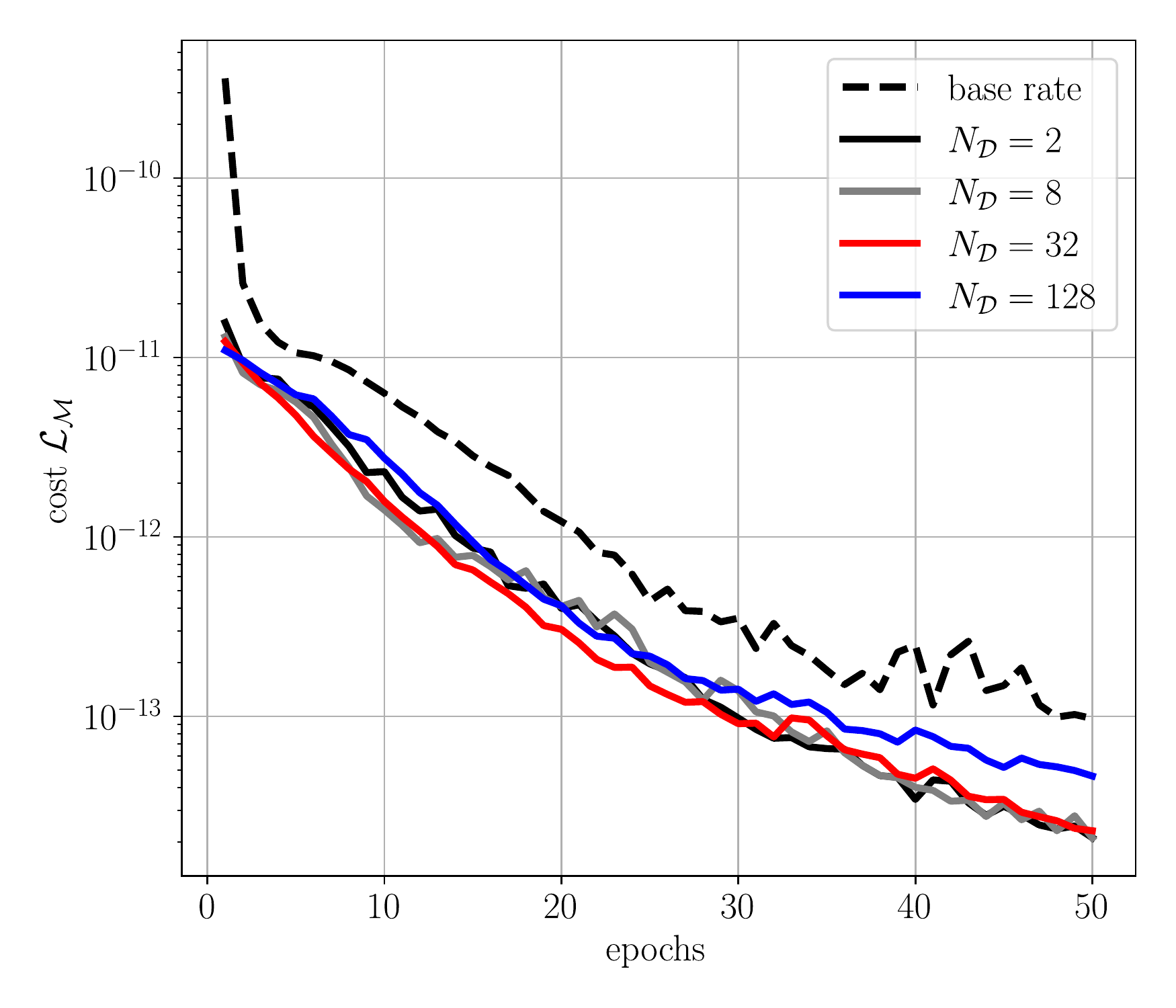}
\caption{FWI cost.}\label{fig:cost}
\end{subfigure} 

\begin{subfigure}[b]{0.41\textwidth}
\centering
\includegraphics[width=\textwidth]{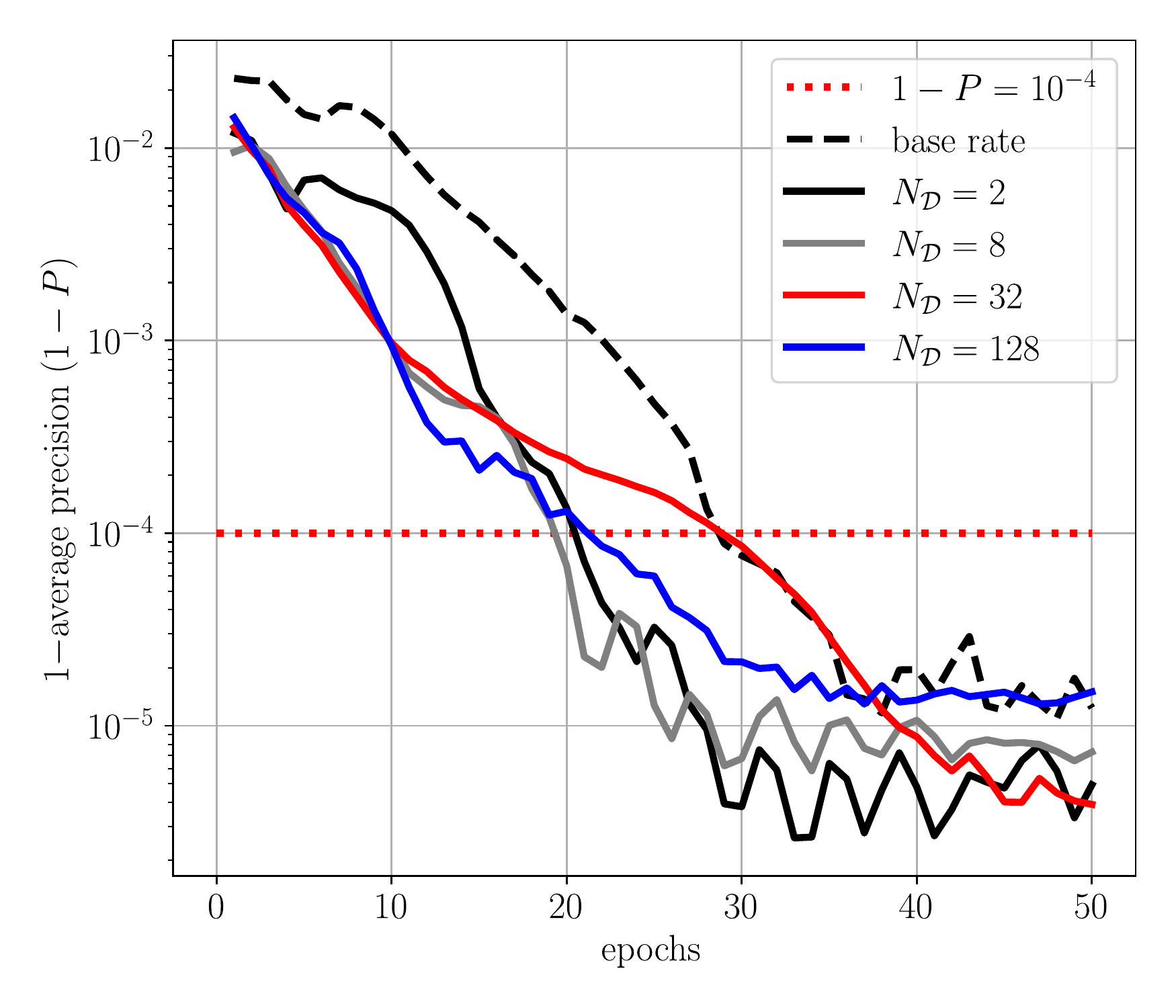}
\caption{Average precision.}\label{fig:avgprecision}
\end{subfigure}
\caption{Influence of using transfer-learning on convergence.}\label{fig:convergence}
\end{figure}

The potential benefit of transfer learning is established with a basic investigation using the ellipsoidal dataset. The convergence behavior is compared in terms of the learning histories throughout the FWI. Both the cost function $\mathcal{L}_\mathcal{D}$ and average precision \cite{zhang_average_2009} w.r.t. the true material distribution $\gamma_\mathcal{M}$ are considered. The average precision $P$ is a preferred metric in object detection and segmentation, which is similar to the task of void detection. See the APPENDIX for a detailed description. Pretrained models with different levels of pretraining are compared to a baserate obtained without pretraining. The models differ in terms of amount of data used during the supervised pretraining, which is $N_\mathcal{D}=2, 8, 32, 128$. As usual in transfer learning, the learning rate has to be decreased in the subsequent training, i.e. the FWI. Interestingly, a larger $N_\mathcal{D}$ leads to a smaller optimal learning rate, obtained by hyperparameter tuning. Additionally some layers of the pretrained network are typically frozen, which in this work was determined empirically to the first 15 layers. The comparison of the learning histories is illustrated in~\cref{fig:convergence}, where the mean over $15$ different inversions is presented.

A suitable recovery of the void with limited artefacts is typically reached when the average precision is larger than $P>1-10^{-4}$, indicated by the red dotted line in~\ref{fig:avgprecision}. The pretrained networks reach this threshold at around 20 epochs, while no pretraining leads to a recovery after 30 epochs. Thus a speed-up of about $1.5$ is achieved. Interestingly, a larger dataset does not seem to yield an additional benefit. This might in later work be alleviated by improved hyperparameters or network architectures.

\begin{figure}[htb]
\centering
\resizebox{0.49\textwidth}{!}{
\begin{tikzpicture}
\node at (0,0) (picA){
    \includegraphics[width=0.3\textwidth]{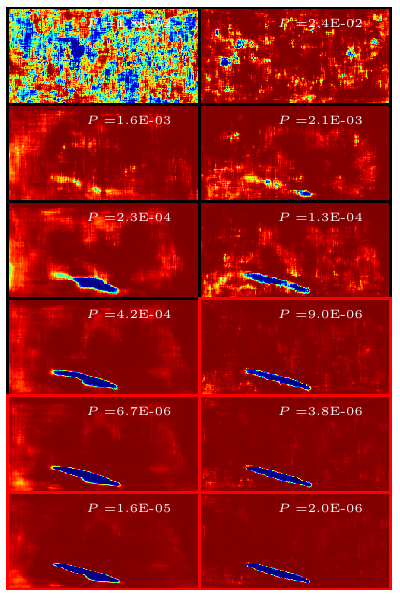}
    };
\node at (-1.4,4.2) {\begin{tabular}{c} No Pretraining\end{tabular}};
\node at (1.3,4.2) {\begin{tabular}{c} Pretraining\end{tabular}};

\node at (-3.5,3.2) {epoch 0};
\node at (-3.5,3.2-1.28) {epoch 4};
\node at (-3.5,3.2-2*1.28) {epoch 8};
\node at (-3.5,3.2-3*1.28) {epoch 12};
\node at (-3.5,3.2-4*1.28) {epoch 16};
\node at (-3.5,3.2-5*1.28) {epoch 20};

\node at (0,-4.5) (picB) {
	\includegraphics[width=0.31\textwidth]{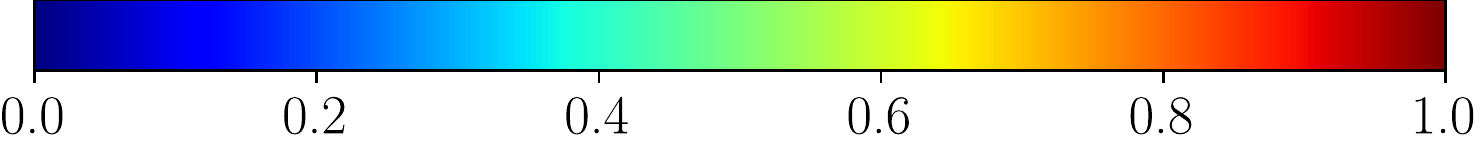}	
};

\end{tikzpicture}
}

\caption{Representative prediction histories starting at epoch 0 until epoch 20. On the left without pretraining and on the right with pretraining using $N_\mathcal{D}=32$ samples. The red boxes highlight the epochs, where the desired threshold of $10^{-4}$ is breached.}\label{fig:gammahistory}
\end{figure}
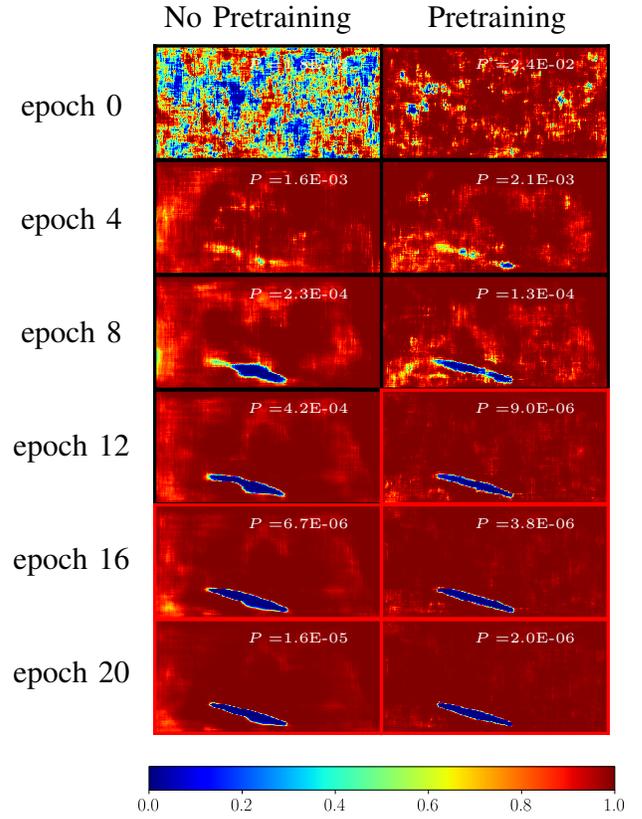

A representative prediction history is illustrated in~\cref{fig:gammahistory} showing the learned state at every 4 epochs starting from epoch 0 until epoch 20. The pretrained model is trained on $N_\mathcal{D}=32$ samples. The red boxes indicate the epochs at which the desired average precision of $P>1-10^{-4}$ is achieved. This shows, that the pretrained model identifies the void about 4 epochs earlier. In later epochs the average precision, depicted in white in the right upper corners, remains significantly lower in the pretrained case. This highlights the additional benefit of using transfer learning when using a neural networks as an Ansatz of the inverse field. 






\section{OUTLOOK}
Although, the obtained speed-up is relatively modest, the results support the proof-of-concept of using transfer learning within a neural network Ansatz in FWI. The lack of improvement of dataset sizes larger than two indicate an unoptimal loss landscape when the problem is transferred from the supervised pretraining to the FWI. This is caused by the rapid change from the data-driven loss function in~\cref{eq:cost_func} to the FWI loss functional from~\cref{eq:measurementloss}. This problem might be alleviated by improved network architectures or hyperparameters, where the level of pretraining, the amount of regularization, e.g. $L^{2}$, the number of frozen layers, and the learning rate after pretraining are the key-parameters to be tuned. An alternative is to avoid the rapid shift in the optimization landscape by pretraining the network not with labeled data, but with multiple FWI cycles with the FWI loss funcitonal~(\cref{eq:measurementloss}). As the understanding of the change in loss landscape is essential in the design of an FWI enhanced by transfer learning, studying its loss landscape with visualization tools such as \cite{li_visualizing_2018} is the key tool.

\section{CONCLUSIONS}
The aim of this work was to establish a proof-of-concept of using transfer learning within the optimization framework of adjoint-based FWI optimization. This was enabled by the neural network Ansatz of the inverse field, previously established as beneficial regularizers \cite{herrmann2023, xu_neural_2019, berg_neural_2021}. A supervised pretraining was employed before the application to the FWI. Although modest speed-ups of only $1.5$ were achieved, the method shows potential. With further investigations into the optimization landscape of FWI with neural networks for the inverse field, significant improvements might be possible w.r.t. the convergence properties.




\section*{APPENDIX} \label{sec:Appendix}

To apply the average precision metric \cite{zhang_average_2009} to the given problem, it has to be considered a classification problem, in this case a binary classification problem, i.e. either a void $\gamma(\boldsymbol{x})=0$ or undamaged material $\gamma(\boldsymbol{x})=1$. This enables the use of true positives $T_P$, false positives $F_P$, true negatives $T_N$ and false negatives $F_N$. The average precision takes both the precision $p$ and the recall $r$ into account. The precision $p$ is the number of true positives over all positives, i.e. true and false positives
\begin{equation}
p=\frac{T_P}{T_P+F_p},
\end{equation} 
while the recall $r$ relates the true positives to the true positives and false negatives.
\begin{equation}
r=\frac{T_P}{T_P+F_N}
\end{equation}
Thus, the precision $p$ represents the level of the false positive rate and recall represents the level of false negative rate. Determining if a value is true (if $\gamma(\boldsymbol{x})\leq t$) or false (if $\gamma(\boldsymbol{x})>t$), depends on a threshhold $t$, which can be varied, such that the precision and recall become functions $p(t)$ and $r(t)$. These can be plotted against each other, yielding the so-called precision-recall curve, which showcases the tradeoff between precision and recall. The average precision is then obtained by integrating the precision-recall curve, which can be approximated by the following summation of discrete threshold intervals $[t-1,t]$.
\begin{equation}
P=\sum_t \frac{R_t-R_{t-1}}{P_t}
\end{equation}

The average precision provides a more accurate estimate of the quality of the prediction compared to the mean squared error, as it is alleviates imbalances in the field, i.e. only small voids in a large intact domain. For implementational details, c.f. to the scikit-learn implementation and documentation \cite{scikit-learn} also used in this work.


\section*{ACKNOWLEDGMENT}

This article is dedicated to Bill and Renate Steen.


\bibliographystyle{IEEEtran}




\end{document}